\documentclass[11pt]{article}

\usepackage[preprint]{coling}

\usepackage{times}
\usepackage{latexsym}

\usepackage[T1]{fontenc}

\usepackage[utf8]{inputenc}

\usepackage{microtype}

\usepackage{inconsolata}

\usepackage{graphicx}
\usepackage{amsmath}
\usepackage{ascmac}

\title{Enhancing Impression Change Prediction in Speed Dating Simulations \\Based on Speakers' Personalities}

\author{
  \textbf{Kazuya Matsuo\textsuperscript{1}},
  \textbf{Yoko Ishii\textsuperscript{1}},
  \textbf{Atsushi Otsuka\textsuperscript{1}},
  \textbf{Ryo Ishii\textsuperscript{1}},
  \textbf{Hiroaki Sugiyama\textsuperscript{1}},
  \\
  \textbf{Masahiro Mizukami\textsuperscript{1}},
  \textbf{Tsunehiro Arimoto\textsuperscript{1}},
  \textbf{Narichika Nomoto\textsuperscript{1}},
\\
  \textbf{Yoshihide Sato\textsuperscript{1}},
  \textbf{Tetsuya Yamaguchi\textsuperscript{1}},
\\
\\
\textsuperscript{1}NTT Corporation
\\
  \small{
    \textbf{Correspondence:} \href{mailto:kazuya.matsuo@ntt.com}{kazuya.matsuo@ntt.com}
  }
}

\begin{document}
\maketitle
\begin{abstract}
  This paper focuses on simulating text dialogues in which impressions between speakers improve during speed dating. 
  This simulation involves selecting an utterance from multiple candidates generated by a text generation model that replicates a specific speaker's utterances, aiming to improve the impression of the speaker. 
  Accurately selecting an utterance that improves the impression is crucial for the simulation. 
  We believe that whether an utterance improves a dialogue partner's impression of the speaker may depend on the personalities of both parties.
  However, recent methods for utterance selection do not consider the impression per utterance or the personalities.
  To address this, we propose a method that predicts whether an utterance improves a partner's impression of the speaker, considering the personalities.
  The evaluation results showed that personalities are useful in predicting impression changes per utterance. 
  Furthermore, we conducted a human evaluation of simulated dialogues using our method.
  The results showed that it could simulate dialogues more favorably received than those selected without considering personalities. 
\end{abstract}

\section{Introduction}\label{sec:intro}
Many people spend significant time and cost on speed dating to find their lifelong partners. 
To mitigate this, a previous research has proposed a method for predicting speed dating results before physically interacting with potential partners\cite{ishii2023prediction}. 
This method enables participants to engage only with pre-selected partners, optimizing their time and resources.

Despite the pre-selection, participants must engage in real dialogues with their potential partners, without opportunities for retries. 
Initial nervousness may lead to poor word choices, thereby risking a negative first impression. 
Consequently, a potentially compatible match might not select them, even if mutual compatibility exists.
If we were able to simulate dialogues that leave favorable impressions (e.g., \textit{love scale} \cite{rubin_loveliking}) on their partners, 
they could understand their partners' preferred dialogue styles (e.g., some are good listeners while others talk more) in advance. 
This can minimize the unfortunate missteps and improve the efficiency of actual interactions. 
However, utterances and dialogues that leave a positive impression may vary depending on speakers' personalities, making such simulation challenging.

\begin{figure}[t]
	\centering
	\includegraphics[scale=0.4]{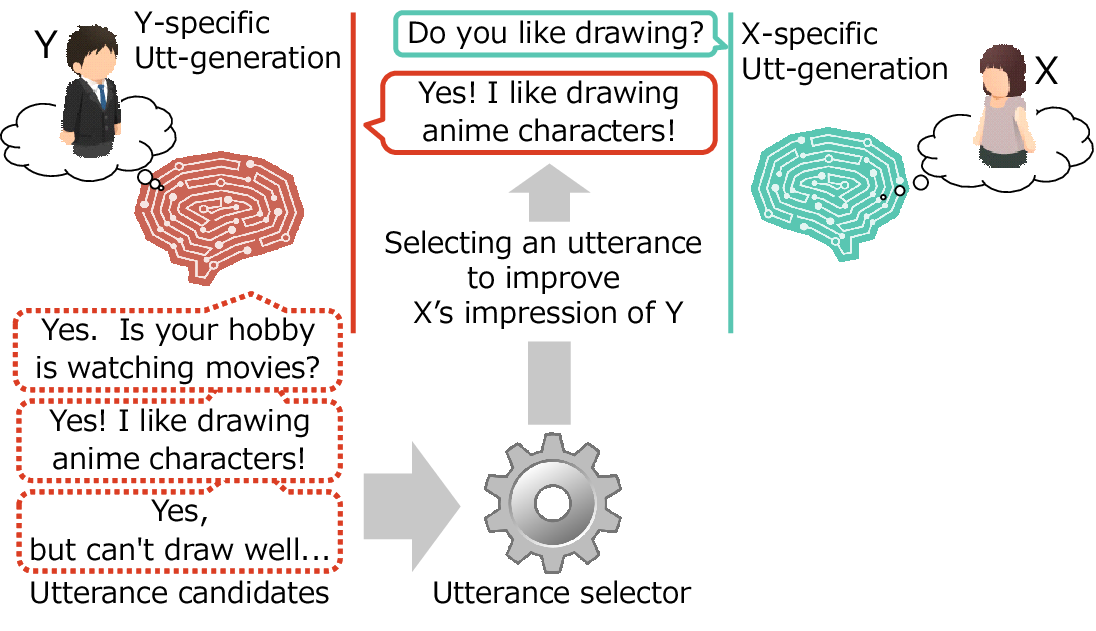}
	\caption{Example of expected dialogue simulation.}
	\label{fig:ex_dialogue_simulation}
\end{figure}

This research aims to realize this dialogue simulation system. 
As shown in Fig. \ref{fig:ex_dialogue_simulation}, this system involves choosing an utterance that would improve speaker X's impression of speaker Y from multiple candidates 
generated by a text generation model capable of reproducing speaker X/Y's utterances. 
In other words, this system requires two key methodologies: 
(1) \textit{person-specific utterance generation}, which replicates what a person is likely to say, 
and (2) \textit{impression-based utterance selection}, which predicts and selects an utterance judged to improve the impression.
Person-specific utterance generation, essential for reproducing dialogues between speakers X and Y, has been explored in various studies \cite{li2016persona,kottur2017persona,higashinaka2018role,mitsuda2022fine,otsuka_personadialogue2024}. 
Simultaneously, to control the dialogue simulation such that the impressions between speakers improve, impression-based utterance selection is also essential. 
However, recent utterance evaluation methods \cite{mehri2022_report_dialogevaluation,lowe2017_auto_turing_test,venkatesh2018evaluating,sugiyama2023empirical,mehri2020usr,mehri2020_FED,zhang2022_investigating_PLM,tsuta2023rethinking} 
do not consider how a target utterance might change a partner's impressions, nor how speakers' personalities influence these changes, 
while they predominantly focus on adequacy, fluency, and coherence. 
Existing impression prediction methods \cite{joel2017romantic,eastwick2022predicting,ishii2023prediction}, while considering personalities, 
overlook dialogue details, making it impossible to estimate impression changes per utterance. 

To address this issue, we propose a novel impression change prediction method. 
Specifically, we build the prediction model using a corpus in which each utterance is annotated with a dialogue partner's impression (i.e., Rubin's love scale) and both parties' personalities. 
This model predicts whether a target utterance in a dialogue context improves a dialogue partner's impression of the speaker based on the personalities and the dialogue context. 
Using this model for utterance selection combined with a person-specific utterance generation, we can simulate dialogues in which the impressions improve.

The contributions of this paper are as follows. 
\begin{itemize}
  \item We propose a novel method that predicts whether an utterance improves a dialogue partner's impression of the speaker, considering both parties' personalities.
  \item We show that the personalities are useful for the impression change prediction. 
  \item We show that we can simulate speed dating dialogues in which the impressions improve, using our method for utterance selection.
\end{itemize}

\section{Corpus}\label{sec:corpus}
To construct the model to estimate whether an utterance improves a dialogue partner's impression of the speaker, 
we annotated a love scale score to each utterance in the Multi-Modal Speed Dating (MMSD) corpus \cite{ishii2023prediction}, which includes dialogues from 625 Japanese pairs. 
Each pair's data is annotated with a 25-item profile and 32 psychological scales (see Appendix \ref{sec:appendix:personality}). 
These scales, reflecting diverse values, are stated to be crucial for expressing one's thoughts on family and relationships, personality, and values in the context of speed dating.
The corpus comprises 15-minute audio tracks with corresponding transcriptions.
To annotate each text utterance with a love scale score, we had 10 of the participants in the MMSD corpus review their dialogues and 
answer 13 love scale-related questions  (see Appendix \ref{sec:appendix:love-scale}) every second, recording scores only when they changed. 
Each of the 13 items was rated on a 9-point Likert scale. 
We assigned each score to the utterance at the time of the change and average the 13 scores per utterance.
Within this corpus, we found 287 utterances that increased, 155 that decreased, and 14,784 that did not change the love scale scores, all conducted by 50 male-female pairs. 
Note that this corpus cannot be made public due to the inclusion of a significant amount of sensitive personal information. 
However, it can be reproduced by following the method described in \cite{ishii2023prediction} and in this section.

\section{Proposed method: Impression change prediction based on speakers' personalities}\label{sec:proposed}
\subsection{Task definition}\label{sbsec:task}
Our task is to predict whether an utterance in a dialogue context increases the love scale score that speaker X feels to Y (hereafter, \textit{the love-score from speaker X to Y}). 
This task can be formulated as :
\begin{equation}
	T = f(P,u,D).
\end{equation}
where $u$ represents a target utterance, $D$ represents the dialogue history up to $u$, 
and $P$ shows both speakers' personalities, essential for predicting the love-score from speaker X to Y after their dialogue \cite{ishii2023prediction}. 
$T$ is the result for estimating changes in the love-score.

\begin{table*}[t]
  \centering
  \small
  \begin{tabular}{l|c|c|c|c}\hline
		Features assigned to target utt. & Acc.($\uparrow$) & Precision($\uparrow$) & Recall($\uparrow$) & $F_1$($\uparrow$) \\\hline\hline
		(a) Only $D$: dialogue history & 0.51 $\pm$ 0.070 & 0.56 $\pm$ 0.089 & 0.73 $\pm$ 0.34 & 0.57 $\pm$ 0.19 \\
		(b) Only $P$: personalities & 0.61 $\pm$ 0.093 & 0.62 $\pm$ 0.081 & 0.85 $\pm$ 0.15 & 0.71 $\pm$ 0.067 \\
    (c) $P$+$D$ & \textbf{0.69 $\pm$ 0.095}${}^\text{a,d}$ & \textbf{0.66 $\pm$ 0.10} & \textbf{0.89 $\pm$ 0.047} & \textbf{0.75 $\pm$ 0.071}${}^\text{d}$ \\\hline
		(d) None (baseline)  & 0.57 $\pm$ 0.047 & 0.58 $\pm$ 0.051 & 0.82 $\pm$ 0.12 & 0.68 $\pm$ 0.044 \\\hline
  \end{tabular}
  \caption{Results of the ablation study. Bold items indicate the highest value for each metric. Superscripts a-d indicate significant differences with methods (a)-(d) respectively ($p<0.05$).}
  \label{table:result_ablation}
\end{table*}

\subsection{Classification model}\label{sbsec:classmodel}
To construct the model to estimate whether an utterance increases the love-score from the partner to the speaker, 
we utilize a pre-trained language model based on a transformer encoder model \cite{vaswani2017attention} and 
apply a fully-connected layer to the output from max-pooling the final hidden states of the input tokens.
We fine-tune this model by using the corpus described in Sec. \ref{sec:corpus}. 
\begin{itembox}[l]{Format of input data}
  \small
  [CLS] Dialogue partner(as another speaker)'s personality [P-SEP] Speaker's personality [SEP] Target utterance [D-SEP] dialogue history [SEP]
\end{itembox}
Here, [CLS], [SEP], [P-SEP], and [D-SEP] are special tokens. 

\section{Experiments}\label{sec:experiments}
In this section, we present the empirical findings of the following research questions guiding our experiments: 
\textbf{RQ1: }\textit{Are speakers' personalities useful in predicting whether an utterance improves a dialogue partner's impression of the speaker?}
\textbf{RQ2: }\textit{Can we simulate dialogues in which partners' impressions improve, using our method for utterance selection?}

\subsection{RQ1: Are personalities useful in predicting impression changes per utterance?}\label{sbsec:auto_ex}
To investigate RQ1, we conducted an ablation study on the classifier described in Sec. \ref{sec:proposed}. 
\subsubsection{Experimental setting}\label{sbsbsec:auto_ex_setting}
\textbf{Fine-tuning training: }
We extracted training data from the corpus described in Sec. \ref{sec:corpus}, aiming to balance instances with increasing love scale scores and those with non-increasing scores. 
The training dataset comprises 549 utterances: 280 with increasing scores and 269 with non-increasing scores, the latter of which includes instances where scores either decreased or remained unchanged.
Each data is annotated with speakers' personalities and 10-utterance dialogue history. 
Using this dataset, we fine-tuned a pre-trained Longformer model \cite{beltagy2020longformer}. 
For cross-validation, 48 pairs were held constant, while the two excluded pairs were evenly divided for validation and testing.
The detail of training setting is described in Appendix \ref{sec:appendix:training_detail}

\noindent\textbf{Comparison methods: }
We believe that how an utterance affects a partner's impression of the speaker relies on their personalities as well as the dialogue context, as described in Sec. \ref{sec:intro}.
To investigate this, we evaluated models trained with personalities ($P$) and dialogue history ($D$), models trained with just one of these features, and models focusing solely on the utterance as a baseline.

\subsubsection{Results}\label{sbsbsec:auto_ex_result}
We utilized Accuracy, Precision, Recall, and $F_1$ score to evaluate the performance. 
The results are shown in Table \ref{table:result_ablation}. 
Classifiers trained with both personalities and dialogue history showed outperformed the baseline in terms of accuracy and $F_1$ score. 
On the other hand, classifiers trained solely with dialogue history showed similar or worse performance compared to the baseline without significant differences
and performed significantly lower accuracy than those trained with both features. 
Moreover, classifiers trained solely with personalities showed similar or better compared to the baseline, without significant differences, 
while they had accuracy comparable to those trained with both features. 

These results show that personalities are useful features for estimating impression changes, 
and while dialogue history alone is less effective, its usefulness increases when combined with personalities.
This is because, while dialogue history is essential for classifiers to understand the context, 
an utterance's effect on improving an impression within that context depends on the personalities.

\subsection{RQ2: Can we simulate dialogues in which impressions improve?}\label{sbsec:human_ex}
To investigate RQ2, we conducted a human evaluation of simulated dialogues involving 20 participants from the MMSD corpus. 
\subsubsection{Experimental setting}\label{sbsbsec:human_ex_setting}
\noindent\textbf{Participant-specific utterance generation: }
We built models that imitate the participants' utterances to simulate their interactions. 
Specifically, we pre-trained a transformer encoder-decoder model based on \textit{PPP dialogue} proposed in \cite{otsuka_personadialogue2024}. 
For fine-tuning, we followed the methodology in \cite{otsuka_personadialogue2024} that utilizes LoRA. 
We used the MMSD corpus as our training data for fine-tuning and customized unique LoRA parameters for each participant in the model.

\noindent\textbf{Comparison methods: }
We used the following three utterance selection methods in dialogue simulation: 

\noindent\textbf{1. Baseline selection} chooses the utterance with the highest cohesion score \cite{sugiyama2021dialogue} from the generated candidates. 
We used this baseline to investigate whether we can simulate dialogues in which impressions improve, not considering changes in impressions per utterance. 

\noindent\textbf{2. $P$+$D$-based selection} uses multiple classifiers trained with $P$+$D$ in the cross-validation 
to more accurately identify which utterances will improve the love-score.
This method chooses the utterance with the most classifiers deemed to improve the love-score. 
If this process cannot single out an utterance, we select from the remaining candidates in accordance with the baseline.

\noindent\textbf{3. $D$-based selection} chooses the utterance in the same way as $P$+$D$-based method using the classifiers trained with $D$. 

\noindent\textbf{Simulated dialogues: }
We created three dialogues between speaker X (each participant of this experiment) and Y (another participant in the MMSD corpus) 
by using the above utterance generation models and utterance selection methods. 
Each dialogue consists of 20 utterances, with the first 10 common utterances to each dialogue and the final 10 utterances varying in accordance as above comparison methods. 
The details of the simulation procedure are shown in Appendix \ref{sec:appendix:simulation_procedure}. 

\noindent\textbf{A/B test: }
We conducted an A/B test by presenting each participant with three pairs of dialogues. 
They were instructed to choose the dialogue in which they had a better impression of speaker Y, assuming the role of speaker X. 
They were also asked to disregard any unnatural elements or hallucinations in the dialogues during their selection.

\subsubsection{Results}\label{sbsbsec:human_ex_result}
Table \ref{table:result_human_eval} shows the results. 
Significantly more participants had a better impression of speaker Y with $P$+$D$-based selection than with $D$-based selection and the baseline dialogue.
However, there was no significant difference between dialogues with $D$-based selection and the baseline.
These findings show that it is difficult to simulate dialogues in which impressions improve when considering only cohesion, neglecting impression changes per utterance. 
Additionally, to simulate dialogues in which impressions improve, considering both personalities and dialogue history is more beneficial than focusing solely on dialogue history.
We provide an example of the simulated dialogue utilized for the A/B test in Appendix \ref{sec:appendix:simulation_result}

\begin{table}[t]
  \centering
  \small
  \begin{tabular}{l|ccc}\hline
		 & $P$+$D$ & $D$ & Baseline \\\hline\hline
    $P$+$D$ & -- & 75\%* & 75\%* \\
		$D$ & 25\% & -- & 65\% \\
		Baseline & 25\% & 35\% &-- \\\hline
  \end{tabular}
  \caption{Results of the A/B test. Each value shows the winning rate of the leftmost listed methods. * indicates a significant difference ($p<0.05$).}
  \label{table:result_human_eval}
\end{table}

\section{Conclusion}\label{sec:conclusion}
In this paper, to realize dialogue simulation in which dialogue partners' impressions improve, 
we propose a method that predicts whether an utterance in a dialogue context improves the impression, considering both parties' personalities.
Our findings addressed the research questions as follows:
\textbf{RQ1: }The performance evaluation results showed that the personalities are useful for predicting the impression change per utterance. 
\textbf{RQ2: }The human evaluation of simulated dialogues using our method showed that our method could simulate dialogues more favorably received than those selected without considering personalities. 

\section*{Limitations}
Our proposed method has the following limitations: 
(1) Due to a lack of training data, our proposed model is unable to predict utterances that decrease the love-score. 
As future work, we plan to expand the training data and retrain the model. 
(2) In our dialogue simulation, users are required to collect a large amount of their personality data. 
As future work, we plan to identify which personalities are key to predicting impression changes.

\section*{Ethical Statement}
\noindent\textbf{Preserving user autonomy: }
While our simulation system can generate potentially undesirable dialogues, our work does not force users to follow them. 
We emphasize the importance of explicitly informing users of this when implementing dialogue simulations in specific domains.

\noindent\textbf{Preventing malicious use: }
We recognize malicious use risks of our system, such as personal information extraction or romance scams. 
To mitigate these, the system must verify identities and avoid simulating sensitive personal information.

\noindent\textbf{Informed consent: }
We fully informed our participants about the study's purpose, procedures, risks, and benefits, and they provided written consent. 
Especially, in the human evaluation, participants are informed that the simulation might generate uncharacteristic utterances, presented as their own.
We also made it clear that they could withdraw their task at any time without penalty.

\noindent\textbf{Ethical considerations: }
The human evaluation has been approved by our Ethics Committee (DT ethics-2023-03).

\bibliography{refer}

\appendix
\section{Detail of personalities}\label{sec:appendix:personality}
We provide an overview of the personalities utilized by our proposed method in Table \ref{table:personality_list}. 

\begin{table*}[t]
  \centering
  \begin{tabular}{c|c}\hline
  Profile & Psychological scale\\\hline\hline
  Age & Rosenberg’s Self Esteem Scale (RSES) \\
  Final Education & Self-Consciousness Scale\\
  Department & Immersion Scale\\
  Residence & Big Five Scale\\
  Hometown & Short Version of Egalitarian Sex Role Attitude Scale\\
  Occupation & Gender Identity Scale\\
  Holiday & Trait Shyness Scale\\
  Annual Income & Self-Monitoring Scale\\
  Family & Clothing Interest Questionnaire\\
  Living Situation & Romantic love attitude Scale\\
  Marital History & Lee's Love Type scale 2nd version (LETS-2)\\
  Smoking & Interpersonal Trust Scale\\
  Alcohol & Family Functioning Scale (FACES III)\\
  About Marriage & Friendship Scale\\
  Personality & Kikuchi's Scale of Social Skills (KiSS-18)\\
  Recent Fad & Value Orientation Scale\\
  Favorite Type & Sense of Leisure Scale\\
  Special Skills & Purpose in Life Scale\\
  Favorite Food & Way of Life Scale\\
  Favorite Movie & Privacy Orientation Scale\\
  Favorite Music & Multidimensional Empathy Scale\\
  How to Spend Holidays & Goal Preference Scale in Friendship Situations\\
  Places to Go on a Date & Affinity Motivation Scale\\
  Topics to Talk About & Loneliness Scale\\
  Self-Introduction & Divorce Feeling Scale\\
  & Friendship Measurement Scale\\
  & Love Image Scale\\
  & Self-concealment Scale\\
  & Communication Skills Scale ENDCOREs\\
  & Daily Life Skills Scale\\
  & Subjective Well-Being Inventory (SUBI)\\
  & Situational Interpersonal Anxiety Scale\\\hline
  \end{tabular}
  \caption{List of personalities.}
  \label{table:personality_list}
\end{table*}

\section{Questions for love scale score}\label{sec:appendix:love-scale}
When obtaining love scale \cite{rubin_loveliking}, we asked the following 13 questions to participants. 
Note that each of the 13 items is rated on a 9-point Likert scale: 1 is "Not True" and 9 is "Definitely True". 
\begin{enumerate}
  \item If [loved one] were feeling badly, my first duty would be to cheer him/her up.
  \item I feel that I can confide in [loved one] about virtually everything.
  \item I find it easy to ignore [loved one]’s faults.
  \item I would do almost anything for [loved one].
  \item I feel very possessive toward [loved one].
  \item If I could never be with [loved one], I would feel miserable.
  \item If I were lonely, my first thought would be to seek [loved one] out.
  \item One of my primary concerns is [loved one]’s welfare.
  \item I would forgive [loved one] for practically anything.
  \item I feel responsible for [loved one]’s well being.
  \item When I am with [loved one], I spend a good deal of time just looking at him/her.
  \item I would greatly enjoy being confided in by [loved one].
  \item It would be hard for me to get along without [loved one].
\end{enumerate}

\section{Detail of training setting}\label{sec:appendix:training_detail}
Using this dataset descrived in Sec. \ref{sbsbsec:auto_ex_setting}, we fine-tuned a pre-trained Longformer model \cite{beltagy2020longformer} with 81M parameters for 15 epochs. 
During fine-tuning, we utilized AdamW with bias correction and a learning rate of $2e^{-5}$ \cite{mosbachstability}, a batch size of 32, and binary cross entropy as a loss function. 
Note that this Longformer model was pre-trained using the corpus listed in Table \ref{table:pretraining_dataset} with the same training parameters as in \cite{beltagy2020longformer}.

\begin{table}[t]
  \centering
  \begin{tabular}{c|c}\hline
		Domain & Paragraphs > 512 chars  \\\hline\hline
		News Crawl & 708,877 \\
    Blog Crawl & 726,490 \\
    Wikipedia & 488,789 \\
    Oshiete goo & 1,595,289 \\
    JParaCrawl & 415 \\
    Corp. Website Crawl & 265,694 \\
    Mainichi News Data & 154 \\
    Common Crawl & 1,827,950 \\\hline
  \end{tabular}
  \caption{Datasets used for pre-training the Longformer, all of which are in Japanese.}
  \label{table:pretraining_dataset}
\end{table} 

\section{Detail of dialogue simulation procedure}\label{sec:appendix:simulation_procedure}
We simulated three dialogues between speaker X (each participant of the human evaluation) 
and Y (another participant in the MMSD corpus) in the following steps: 
\begin{enumerate} 
  \item (Preparation) We manually set the initial utterance, ``What is your hobby?'' 
  \item Using X's LoRA and the input utterance, we generate 20 replies. 
  \item An utterance is chosen using the baseline. 
  \item The same process is repeated with Y's LoRA. 
  \item Steps 2-4 are repeated until 10 utterances are generated. 
  \item We create three patterns of the final 10 utterances. 
  For each pattern, Steps 2-5 are repeated with a different method for Step 3.
\end{enumerate}

\section{Example of dialogue simulation result}\label{sec:appendix:simulation_result}
Table \ref{table:ex_result_sim} presents the dialogues simulated: 
condition A using $P$+$D$-based selection and condition B using $D$-based selection.
In the A/B test, the participant chose dialogue A, stating that Y in A seemed more empathetic, unlike the self-centered Y in B.
By considering personalities like hers in utterance selection, our method can enhance the accuracy of simulating dialogues in which impressions improve.
\begin{table*}[t]
  \centering
  \begin{tabular}{c|p{9.5cm}}\hline
		 & Simulated dialogues  \\\hline\hline
		& Y: What is your hobby?\\
		& X: I like to travel. I enjoy anime, manga, and video games!\\
		& Y: I live it, too. I've been doing a lot of domestic these days too, but I do go abroad a lot!\\
		& X: Oh, I see! Where did you go abroad?\\
		& Y: Recently I went to Korea and Taiwan. \\
		The first 10 utterances & X: I have never been to Korea. Taiwan, I want to go there too!\\
		& Y: I recommend Taiwan. \\
		& X: What is good food in Taiwan?\\
		& Y: After all, shrimp is famous. Also, Taiwanese ramen noodles are delicious!\\
		& X: Shrimp and Taiwanese Ramen! What do they taste like?\\\hline
		& Y: The shrimp has a bit unique and delicious taste. \\
		& X: Is it unique? What does it taste like?\\
		& Y: Yes, it is unique. The soup is a bit sweet and spicy, but I am sure it is delicious!\\
		& X: It is a sweet soup. But spicy means it tastes a bit strong, right?\\
		The last 10 utterances & Y: Yes, it is a strong tasted ramen. But it depends on the restaurant!\\
		($P$+$D$-based selection, condition A) & X: I see! Eating out is delicious, isn't it?\\
		& Y: Yes, it is. I've been eating out recently at a restaurant called "Mr. Ramen shop". \\
		& X: "Mr. Ramen shop"? What kind of ramen do we eat? \\
		& Y: Mr. Ramen shop is very popular and it's soup is so rich!\\\hline
		& Y: The shrimp are very juicy and delicious. \\
		& X: So it's shrimp! Where can I get some?\\
		& Y: You can eat at an izakaya. I recommend the shrimp and miso soup!\\
		& X: I see. The shrimp and miso soup are the best. \\
		& Y: I like sushi, have you ever had it in Tokyo?\\
		The last 10 utterances & X: Yes, I have. Recently, a restaurant called "Sushi Restaurant Tokyo" in Asakusa was delicious!\\
		($D$-based selection, condition B) & Y: Ah, Sushi Restaurant Tokyo. There are so many good restaurants!\\
		& X: Yes. Also, I recently visited a popular restaurant called Sushi Restaurant Akihabara!\\
		& Y: I see, I'd like to go there sometime too!\\\hline
  \end{tabular}
  \caption{A pair of simulated dialogues in the A/B test. }
  \label{table:ex_result_sim}
\end{table*}

\end{document}